\documentclass{article}

\usepackage[preprint]{neurips_2023}

\usepackage[utf8]{inputenc} 
\usepackage[T1]{fontenc}    
\usepackage{hyperref}       
\usepackage{url}            
\usepackage{booktabs}       
\usepackage{amsfonts}       
\usepackage{nicefrac}       
\usepackage{microtype}      
\usepackage{xcolor}         
\usepackage{graphicx} 

\title{Large Knowledge Model: Perspectives and Challenges}

\author{%
	Huajun Chen \\
	College of Computer Science, Zhejiang University, Hangzhou, 310027, China \\
	\\
	\texttt{huajunsir@zju.edu.cn} \\  
}

\begin{document}

	\maketitle

	\begin{abstract}
Humankind's understanding of the world is fundamentally linked to our perception and cognition, with \emph{human languages} serving as one of the major carriers of \emph{world knowledge}. In this vein, \emph{Large Language Models} (LLMs) like ChatGPT epitomize the pre-training of extensive, sequence-based world knowledge into neural networks, facilitating the processing and manipulation of this knowledge in a parametric space. This article explores large models through the lens of ``knowledge''. We initially investigate the role of symbolic knowledge such as Knowledge Graphs (KGs) in enhancing LLMs, covering aspects like knowledge-augmented language model, structure-inducing pre-training, knowledgeable prompts, structured CoT, knowledge editing, semantic tools for LLM and knowledgeable AI agents. Subsequently, we examine how LLMs can boost traditional symbolic knowledge bases, encompassing aspects like using LLM as KG builder and controller, structured knowledge pretraining, and LLM-enhanced symbolic reasoning. Considering the intricate nature of human knowledge, we advocate for the creation of \emph{Large Knowledge Models} (LKM), specifically engineered to  manage diversified spectrum of knowledge structures. This promising undertaking would entail several key challenges, such as disentangling knowledge base from language models, cognitive alignment with human knowledge, integration of perception and cognition, and building large commonsense models for interacting with physical world, among others. We finally propose a five-``A'' principle to distinguish the concept of LKM.
\end{abstract}

\section{Language vs Knowledge}
	
\subsection{Human Language and World Knowledge}

Humankind accumulates knowledge about the world in the processes of perceiving the world, with natural languages as the primary carrier of \emph{world knowledge} \cite{Yule}. Historically, the vast majority of world knowledge has been described, documented, and passed on through natural languages. In addition to natural languages that record common-sense knowledge, human has also invented various scientific languages for describing specialized scientific knowledge, for examples, the mathematical languages that describe mathematical models \cite{VerifyStepByStep,MATHDataset}, chemical languages such as SMILE for describing molecular structures \cite{SMILE,Fingerprint}, and genetic languages to model compositions of living organisms \cite{GeneOntology,GOResource,OntoProtein}.
	
However, natural languages merely encode world knowledge through sequences of words, while human cognitive processes extend far beyond simple word sequences. Therefore, since early inceptions of AI, it has been a fundamental objective to explore machine-friendly formats for \emph{Knowledge Representations} (KR) \cite{WhatIsKR}. Typical examples of these endeavors encompass Description Logics \cite{IntroDL} for representing ontological knowledge, Prolog for rule-based logic \cite{prolog2022}, Semantic Networks for depicting conceptual relationships \cite{woods1975whatsInALink}, and various others. 
	
In traditional symbolic AI research, the logical structure and expressiveness of a KR is pivotal to the reasoning ability of the inference machines \cite{WhatIsKR,IntroDL}. Natural language sequences composed of simple words or concepts are generally considered to be unfavorable for machines to make inference, while hierarchical taxonomies, complex ontologies, and rule-based logic are more conducive to machine reasoning. As highlighted later, even in the era of large models, there persists a positive correlation between the complexity level of data representations and the reasoning proficiency of large models.
	
\subsection{Language Models vs Knowledge Graphs}
	
\emph{Knowledge Graphs} (KGs) model world entities, mental concepts and their relationships in form of graph structures, which are typically extracted or derived from natural language descriptions. With many inspirations from traditional symbolic AI, KGs integrate natural language descriptions and structure knowledge. Typical structural forms include: hierarchical structures(e.g., concept graphs), relational structures(e.g., relational entity graphs), temporal structures(e.g.,  event logic graphs), etc. Akin to natural languages, KGs encode knowledge symbolically, ensuring robust reliability, traceable reasoning, and human-centric interpretability.
\begin{figure}[htbp]
    \centering
    \includegraphics[width=1\textwidth]{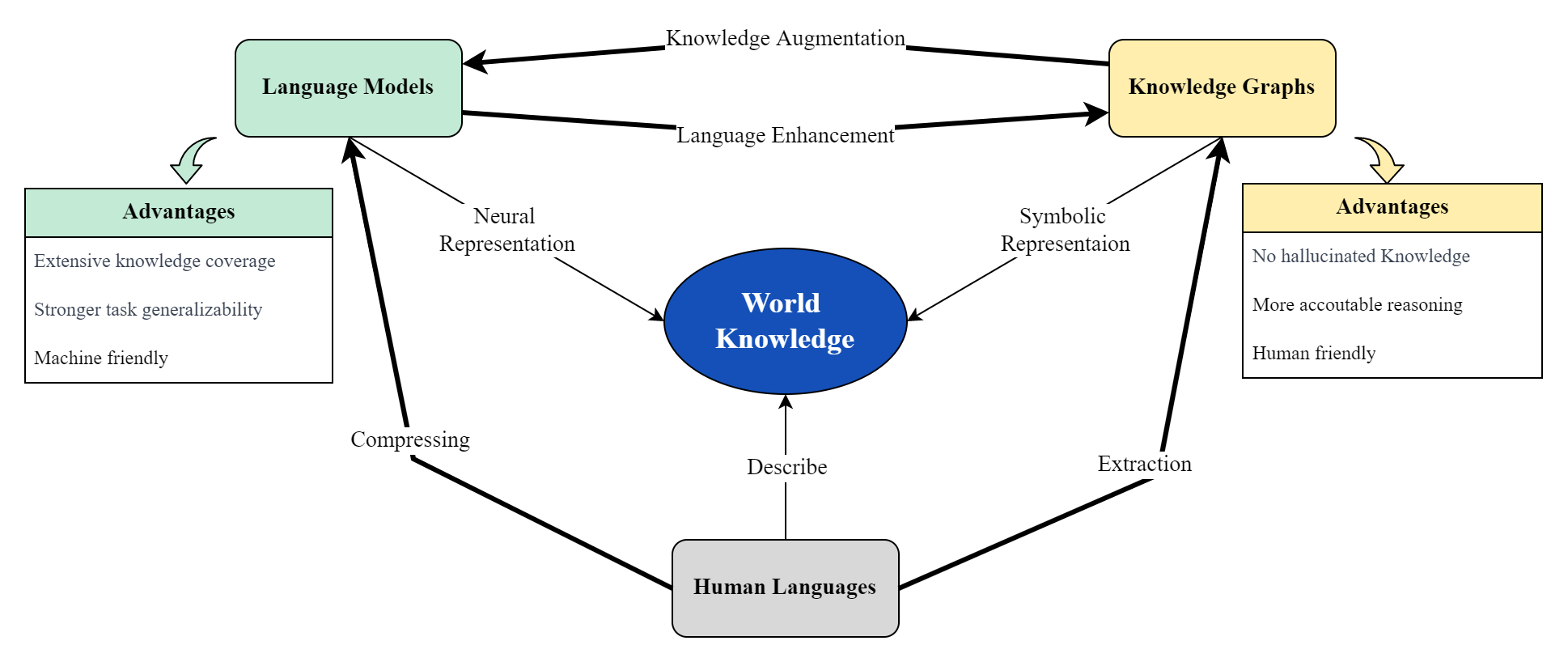}
    \caption{Language, Knowledge, Language Models and Knowledge Graphs.}
    \label{fig:kgllm}
\end{figure}
 
\emph{Large Language Models} (LLMs) \cite{SurveyOfLLM,HarnessLLM}, such as ChatGPT, mark a paradigm shift in knowledge representation and processing. By autonomously compressing or projecting vast textual knowledge in form of human languages into neural networks, LLMs achieve a parameterized representation and handling of world knowledge. LLMs internalize deep word patterns and  interactions from extensive text corpora, establishing a wider coverage of knowledge than conventional symbolic knowledge base. Advanced techniques like prompt or instruction tuning further imbue LLMs with prior human knowledge, expanding their generalization capacity and adaptability to new tasks. Unlike natural language and KGs, LLMs operate on a fully parameterized, machine-friendly basis, albeit less interpretable to humans.
	
As depicted in Figure \ref{fig:kgllm}, both LLMs and KGs are specifically developed for the representation and manipulation of knowledge. KGs provide highly accurate and interconnected knowledge, enabling controlled reasoning and enhanced human readability. Conversely,LLMs offer expansive coverage of knowledge, exhibit enhanced task generalization capabilities, and utilize a neural representation that optimizes machine efficiency. Yet, KGs face scalability challenges, are limited in their ability to extrapolate, and struggle with reasoning transferability. LLMs, despite their power, incur significant training costs, struggle with deep logical reasoning, and are susceptible to hallucination errors.

\begin{figure}[h]
\centering
\includegraphics[width=0.95\textwidth]{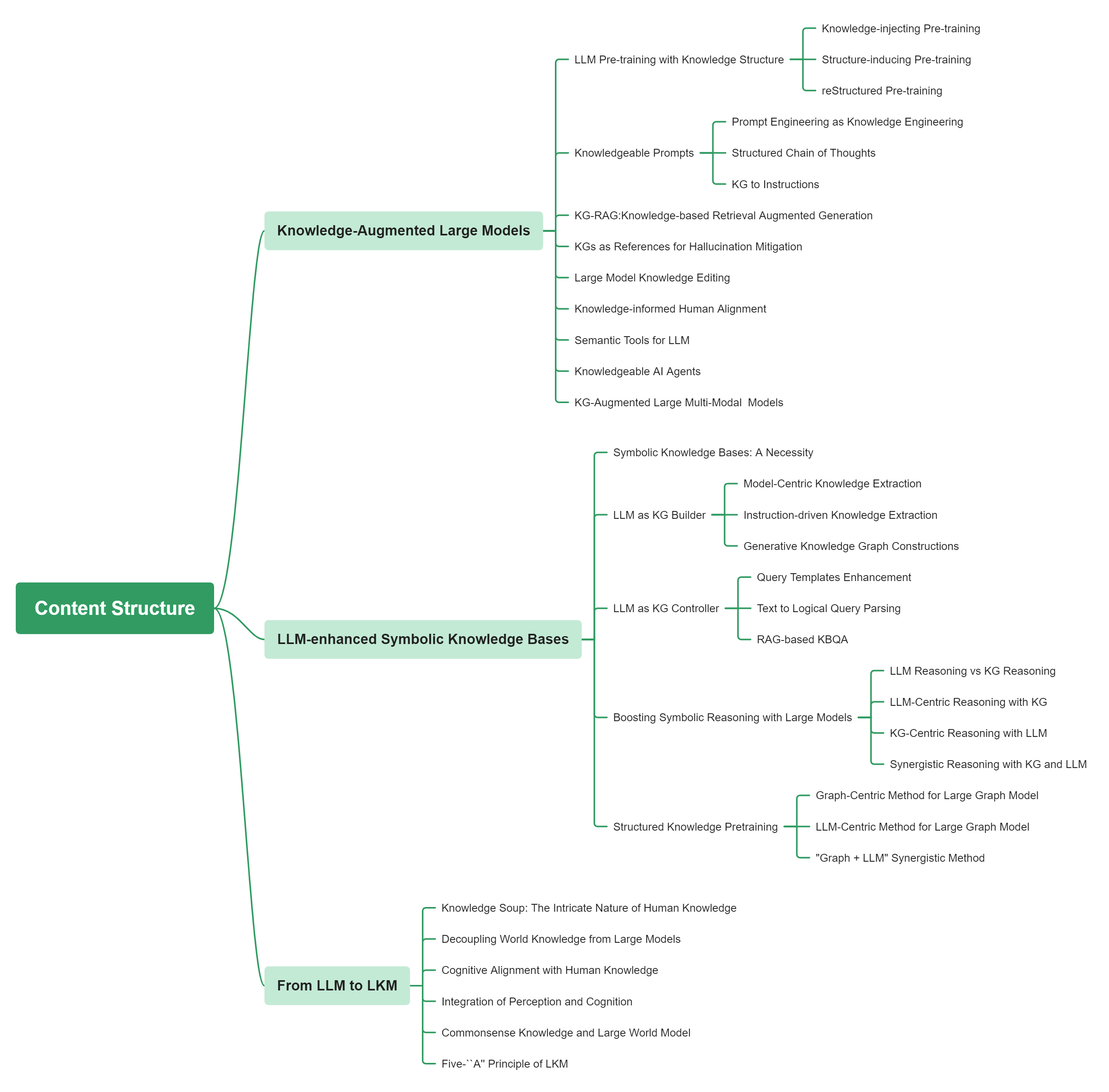}
\caption{An Outline of the Whole Content Structure}
\label{fig:ContentStructure}
\end{figure}

In this article, we aim to investigate large models through the lens of ``knowledge''. Our first focus is on how KGs can enhance LLMs. Topics include knowledge-augmented language modeling, structure-inducing pre-training, knowledgeable prompts, structured Chains of Thought (CoT), semantic tools for LLMs, editing knowledge in large models, and the development of knowledgeable AI agents. We further examine the role of LLMs in enriching traditional symbolic knowledge bases, encompassing the utilizing of LLMs as KG builders and controllers, structured knowledge pre-training, LLM-driven symbolic reasoning. Acknowledging the intricate nature of human knowledge, we then conceptualize a \emph{Large Knowledge Model} (LKM) in purpose to handle the diversity of structures of human knowledge. This may involve approaches of decoupling knowledge from language models, restructure pretraining with structure knowledge, building large commonsense model, among others. To distinguish the concept of LKM, we propose a five-``A'' principle from the aspects of \textbf{A}ugmented pretraing, \textbf{A}uthentic knowledge, \textbf{A}ccountable reasoning, \textbf{A}bundant knowledge coverage, and \textbf{A}ligned with knowledge.

\section{Knowledge-Augmented Large Models}
	
The first step towards \emph{Large Knowledge Models} is to augment LLM with knowledge. In this context, ``knowledge'' specifically refers to more standardized semantic representations such as domain terminology or ontologies, more structured knowledge representations such as a KG, or richer logic descriptions such as  text with CoT. We will discuss how knowledge can be effectively applied in various aspects in the life cycle of a LLM, including pre-training, fine-tuning with instructions, prompt engineering, Chain-of-Thought, AI agents, among others.

\subsection{Expressiveness vs. Scaling Law: The Dichotomy in Knowledge Representation}

Traditional research in symbolic Knowledge Representation (KR) mainly focuses on the link between representation expressiveness and reasoning ability, suggesting that more expressive or complex representations facilitate stronger reasoning. However, this perspective often overlooks a vital factor: the scale of knowledge also significantly influences reasoning. For example, the \emph{scaling law} in large models \cite{kaplan2020scaling} indicates that diversified reasoning abilities only emerge when the model reaches a certain size. This phenomenon can be attributed to the fact that an increase in model size results in an increase of knowledge coverage, which in turn enables more sophisticated reasoning tasks.

\begin{figure}
    \centering
    \includegraphics[width=0.95\textwidth]{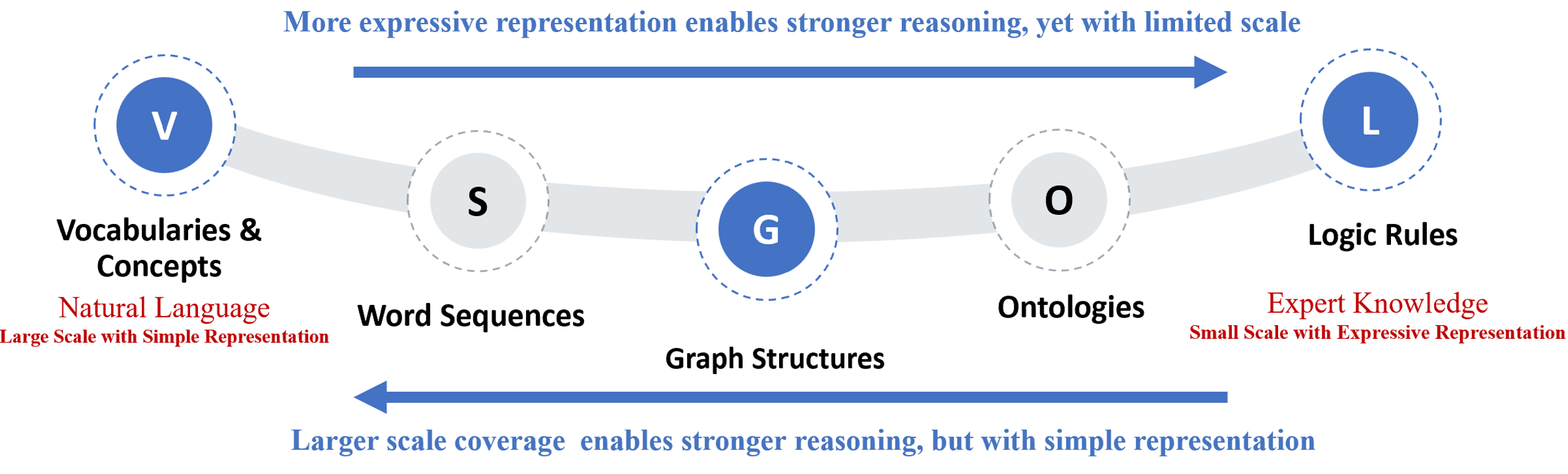}
    \caption{The Dichotomy in Knowledge Representation}
    \label{fig:dichotomy-in-kr}
\end{figure}

Indeed, both expressiveness and scale act as balancing factors for reasoning capabilities as illustrated in Figure \ref{fig:dichotomy-in-kr}. More intricate representations result in more precise knowledge, like an ontology for KGs, thus enabling stronger reasoning, but incuring more challenging to acquire the knowledge in large scale. In contrast, simpler representations, like textual sequences for training large models, facilitate extensive coverage of knowledge but compromise delicate reasoning due to reduced precision.  As we will delve deeper, this inherent contradiction forms the basis of the complementary nature of KGs and large models, driving advancements in augmenting large model with conventional symbolic KR. 
		
\subsection{LLM Pre-training with Knowledge Structure}

Let's first address how to enhance the pre-training stage with various forms of knowledge.

\subsubsection{Code Structure and Thought Logic}

Numerous studies suggest that enhancing corpus structure or incorporating more logic into the training corpus can improve the reasoning performance of large models \cite{ToG,MindMap,InstructProtein,CoK,UnifyKGLLM,ClinGen}. For examples, code languages, characterized by their structured nature and abundant computational logic, would be more conducive to activating the reasoning capabilities of large models compared to natural languages \cite{PoT,PoT-CIRS}. Similarly, Chain-of-Thought (CoT) prompts  \cite{CoT,CoT-SC} contain more logical descriptions than ordinary text, thus being  more effective in stimulating model reasoning ability. Further research reveals that CoT prompts are more effective when the model scale to a certain size, which can be explained as that a larger model contains more knowledge for reasoning, and the activation of these knowledge requires more logical prompts.
	
To delve deeper into the correlations between code structure and model reasoning, we introduce a study about ``Reasoning with Program of Thoughts'' \cite{PoT-CIRS}. This research begins by establishing metrics for quantifying the complexity of code structure and code logic. For instance, codes featuring loop operations are considered as having richer computational logic and assigned with higher scores compared to those with mere variable assignment operations. Subsequently, we compile a series of code prompts with different levels of structure or logic scores, and assess their impact on model reasoning performance. Experimental analyses on various benchmarks reveal that the code prompts with higher structure or logic scores are more conducive to improving model reasoning performance.
	
Although the representation of code structure and CoT logic remains at a relatively simple level, it has nonetheless contributed to enhancing the reasoning capabilities of large models. It can be anticipated that the utilization of data representations with deeper logic and more complex structure, such as knowledge graphs, can further enhance the capabilities of models. 
	
\subsubsection{Knowledge-injecting Pre-training}

\begin{figure}[h]
\centering
\includegraphics[width=1.0\textwidth]{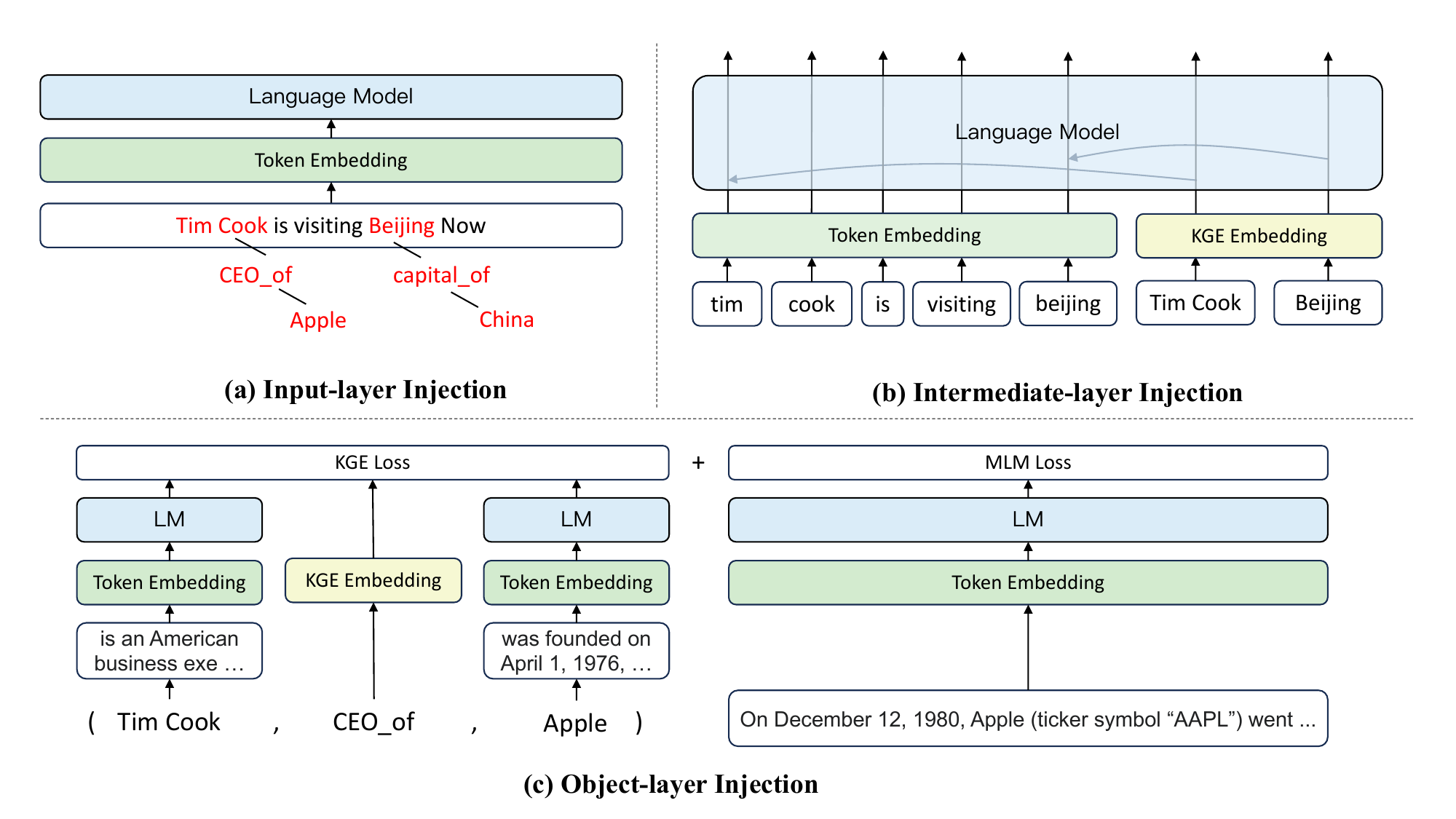}
\caption{Knowledge Injection on Different Layers.}
\label{fig:inject}
\end{figure}

Large models build their reasoning prowess by understanding sequences of words. To achieve deep reasoning similar to human cognitive processes, it's essential to progress beyond basic word semantics to the formation and comprehension of advanced concepts, entities, and their interconnections. Injecting knowledge graph with pre-training models is a promising strategy to enable such type of richer understanding and move beyond the limitations of simple word co-occurrences to embrace the complexity of higher-order reasoning. There exist several strategies for integrating knowledge graphs into pre-trained language models, broadly categorized based on the stage of integration: input layer, architecture layer, and objective layer as illustrated in Figure \ref{fig:inject}.

\begin{itemize}
    \item \emph{Input-layer Injection}. At the input layer, the simplest strategy is to convert knowledge graphs into natural language descriptions, thus expanding the model's input without altering its architecture or training process. Notable examples include K-BERT \cite{kbert} and CoLAKE \cite{CoLAKE}, among others. The core idea revolves around leveraging the relationships between entities in knowledge graphs to enrich the contextual information of entities within sentences. Typically, the process begins with establishing connections between entities in sentences and those in the knowledge graph using entity linking technology. Subsequently, sentences are expanded with triples corresponding to these knowledge graph entities before being fed into the large model.  Injecting knowledge at the input layer fundamentally serves as a form of data augmentation. These methods do not require modifications to the model's architecture and are compatible with various off-the-shelf models, making them easy to implement. Additionally, because both structured and textual knowledge are trained within the same model, this approach avoids mismatches between the textual representation space and the structured knowledge representation space. However, converting inherently structured graphs into text sequences inevitably leads to some loss of semantic representation, and the utilization of structural signals is not fully realized.
    
    \item \emph{Intermediate-layer Injection}. The second strategy involves initially employing structured knowledge representation learning methods, such as KG Embedding, to obtain vector representations of entities and relationships, which are then integrated into language models. Notable examples include ERNIE \cite{ERNIE}, KnowBERT \cite{KnowBERT}, KG-BART \cite{KG-BART}, KT-NET \cite{KT-NET}, and BERT-MK \cite{Bert-MK}. For instance, ERNIE starts by pre-training entity representations using TransE. The original textual input remains unchanged, following alignment and interaction with these entity vectors at the intermediate layer of the model with a simple attention mechanism. Since the entity vector representations have already learned the structured features from the original knowledge graphs, this approach better preserves the graph's structural semantics. However, since textual semantics and graph structure representations are derived from two different pre-training models, there might be interference between textual semantic signals and knowledge structure signals, potentially acting as noise to each other. 

    \item  \emph{Objective-layer Injection}. The third strategy employs knowledge injection at the objective level, often through multi-task learning and refined LOSS optimization. Notable examples include KEPLER\cite{KEPLER}, WKLM \cite{WKLM}, JAKET \cite{JAKET}, and BERT-MK \cite{Bert-MK}. For instance, KEPLER adds a KG embedding Loss on top of the standard masked prediction Loss, using a multi-task learning framework for simultaneous training. As the same encoder is used for both text and entity information, and a single Loss governs the learning of both KG embedding and language model pre-training, this method preserves more structural signals compared to input layer injection and avoid the need for explicit alignment between text and KG  required by architecture layer injection.
\end{itemize}
	
\subsubsection{Structure-inducing Pre-training}

We have already mentioned that human knowledge is not a simple sequence of texts, but rather has a rich structure. In a study called ``Structure-Inducing Pretraining'' \cite{StructIndPretrain}, the role of imposing relational structure in pretrained language models is investigated. The basic ideas involve crafting connections, either at \emph{inter-sample} or \emph{intra-sample} level, to enforce additional structural constraints on the geometric relationships within or between training samples. These constraints range from shallow, implemented via auxiliary pretraining objectives, to more intricate, implicitly integrated through strategies like data augmentation or sample-level contrastive learning.
	
A typical example of \emph{inter-sample} structure constraints manifests in the injection of a protein-protein interaction network into the pretraining of a protein language model. This injection infuses the model with sample-level structural insights, i.e. the relational structures of protein interactions. A typical example of \emph{intra-sample} structure manifests in the linking of entities within a sentence to an external knowledge graph, thereby introducing correlation information in a KG into the training sentences. Both theoretical analysis and experimental verifications are performed and indicate a consistent improvement with structure-inducing across a variety of benchmarks.

\begin{figure}[h]
    \centering
    \includegraphics[width=1.0\textwidth]{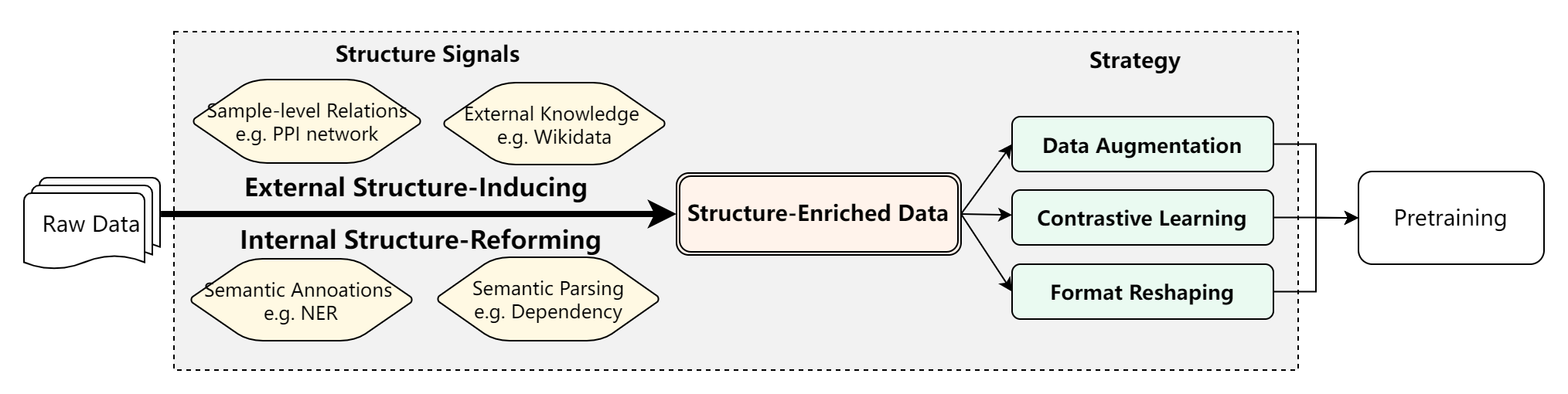}
    \caption{Structure-Augmented Pre-training}
    \label{fig:structaug}
\end{figure}

\subsubsection{reStructured Pre-training}

Another relevant research is so-called reStructured Pre-training \cite{yuan2022restructured}. Rather than imposing external structural knowledge, as in those structure-inducing methods, reStructure approach delves into the structural signals inherent in the data itself. The world's data is rich with various structural signals in different formats. For instance, it could be as straightforward as the ``next token'' signal in a sentence, metadata for a given text, factual knowledge in a triple format, entity or relation annotations, dependency structures between words, and numerous others. To make LLMs better educated, it is justifiable to mine these structure signals automatically or semi-automatically, and restructure all of them into unified  forms for pre-training models.

\subsection{Knowledgeable Prompts}
Next, we will explore how to utilize knowledge to enhance prompt instructions.
\subsubsection{Prompt Engineering as Knowledge Engineering}
Prompts refer to natural language text describing the instructions that an AI should perform, the context of instructions that an AI can inquire of, or the guidance to think that an AI can follow \cite{InstructGPT,CoT,FLAN}.   It has been proven by numerous practices \cite{Auto-CoT,APE,AutoPrompt,KNNPrompt} that the quality of prompts has a huge impact on model outputs. By providing prompts or instructions with more specific details, the model can more accurately comprehend human intentions, leading to outputs that are more aligned with human instructions. 
	
Prompt engineering is essentially comparable to knowledge engineering. Increasing number of research indicates that the complexity of prompts is closely correlated to model reasoning ability. The evolution of prompts ranges from mere textual prompt, CoT prompt with more logic descriptions to more structured prompts such as \emph{Program of Thoughts} \cite{PoT}, \emph{Tree of Thoughts}\cite{ToT}, and \emph{Graph of Thoughts} \cite{GoT}, and even directly using logical rules as prompts \cite{PTR}. As the complexity and expressiveness of prompt representations increases, an improvement of model reasoning ability can be observed. However, there persist a trade-off between richness of prompt representations and scalability of prompts acquisition: higher expressiveness in prompts typically entail increased difficulty for acquisition, which is essentially similar to traditional knowledge engineering. High-quality prompt engineering is also time-consuming and laborious as like knowledge engineering.
	
\subsubsection{Structured Chain of Thoughts}
	
Chain of Thought (CoT) is a special type of prompts describing the thinking procedures that can guide large models to imitate human logical thinking. As previously discussed, employing more intricately structured CoT representations, such as Tree of Thoughts \cite{ToT} or Graph of Thoughts \cite{GoT}, would bolsters the model's capacity for reasoning. 
	
Moreover, KG is obviously conducive to enhancing the structure of CoT prompts. On the one hand, The structured knowledge in a KG proves useful in the formulation and generation of structured CoT prompts \cite{CoK,ToG,MindMap}, to form a CoT knowledge graph (KGoT). On the other hand,  through the entity linking technologies, structure correlations among concepts and entities in the KG can be used to expand and enhance CoT in natural language forms, making the textual CoT more in line with human thinking patterns based on associations,thereby enhancing their effectiveness in guiding AI reasoning.
	
\subsubsection{KG to Instructions}

Instruction tuning \cite{InstructGPT,FLAN} refers to further training base models with task-specific instructions, typically converted from task-specific training data. Since data from various tasks are converted to an unified input form as instructions or prompts, instruction-tuning with massive downstream task would significantly improve model's generalization ability, that is, the ability to solve new problems or to complete new tasks.
	
Knowledge Graphs serve a dual roles of enhancing instruction-tuning. On the first hand, KGs can be directly injected into instruction prompts. For instances, KnowPrompt \cite{KnowPrompt} utilizes the relational data in a KG to bolster the contextual depth of instruction prompts. Similarly, KANO \cite{KANO} integrates structured knowledge from a chemical KG, sourced from Wikipedia, to refine functional prompts, thereby improving the learning process for molecular pretraining. 
	
\begin{figure}[h]
    \centering
    \includegraphics[width=0.95\textwidth]{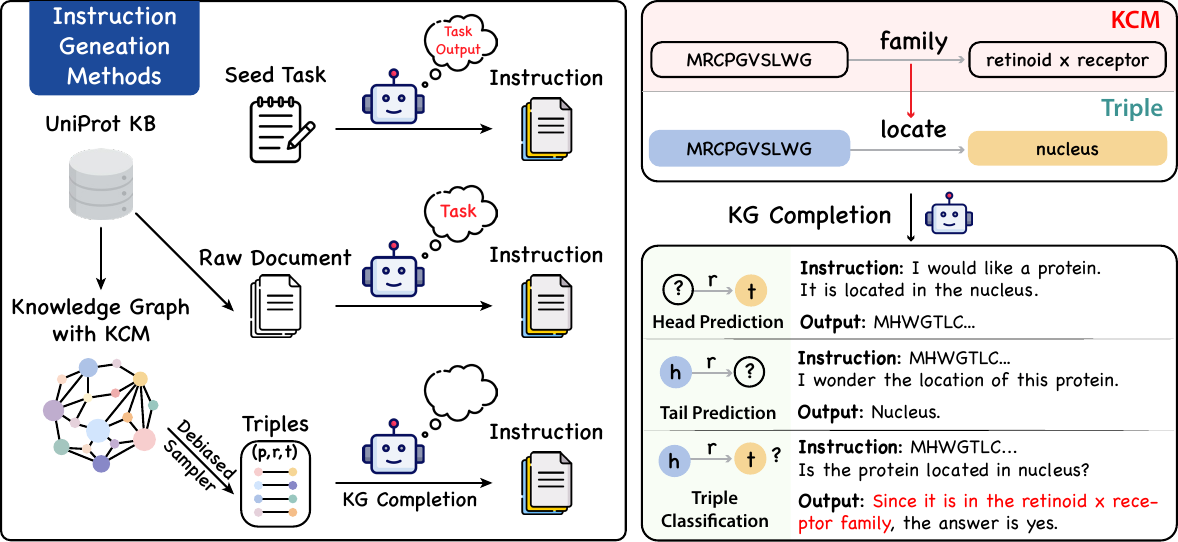}
    \caption{InstructProtein: an example of building instruction dataset with a KG \cite{InstructProtein}}
    \label{fig:kg2ins}
\end{figure}

On the other hand, KGs can be directly converted to instructional datasets. As exemplified by DeepKE-LLM \cite{knowlm}, a KG refined from WikiData is employed to automatically generate vast amount of training instructions for knowledge extraction. Essentially, these instructions converted from the KG represent a novel form of distant supervision, which leverages the structured information in KGs to generate instructions that guide models in understanding and extracting relevant information across diverse contexts. As shown in Figure \ref{fig:kg2ins}, InstructProtein \cite{wang2023instructprotein} introduces a knowledge graph-based instruction generation framework to construct a high-quality instruction datasets for training a protein language model . In particular, the instructions inherit the structural relations between proteins and function annotations in a protein knowledge graphs, which empowers the model to engage in the causal modeling of protein functions, akin to the chain-of-thought processes in natural languages.
	
Compared to free text, a KG is data corpus with richer logical structures and higher knowledge density. Many fields of knowledge graph data come from long-term artificial accumulation (such as Gene Ontology) or natural structured data transformation. Fully mining and using the knowledge structure contained in the knowledge graph (such as hierarchical concepts, entity relationship, temporal causality, rule logic, etc.) to enhance and enrich the prompt instruction or assist in the construction of more logical instruction data set, can help the model to complete more complex tasks.
	
\subsection{KG-RAG: Knowledge-based Retrieval Augmented Generation }

Retrieval-Augmented Generation (RAG) refers to the enhancement of large models' generative capabilities through querying an external knowledge base, which could be external search engines,  domain-specific corpora, structured databases, or knowledge graphs. Due to its cost-effectiveness in training, avoidance of model hallucinations, and compliance with privacy and copyright standards, RAG has become an essential tool in large model applications.

Knowledge-augmented RAG leverages the conceptual hierarchies, entity connections, and logical rules within knowledge graphs to improve retrieval processes.   The essence of RAG involves segmenting documents or images into data chunks, which are embedded and stored in a vector database for retrieval. The primary issue with chunking lies in the risk of losing the document's inherent logic and contextual relationships. Vector representations of these chunks offer merely a basic feature vector for similarity computation, falling short of capturing the deeper semantic connections among entities or concepts—a gap knowledge graphs were designed to bridge.

A practical way to apply KG-RAG is through the creation of entity or concept trees, enhancing RAG into versions like Entity-RAG or Topic-RAG. Using knowledge graphs to build hierarchical structures of concepts, topics, or entities, and connecting these elements within document chunks to the knowledge graphs, can significantly improve the retrieval process through these organized knowledge hierarchies. A typical example is Tree-RAG \cite{Tree-RAG}, which merges RAG search with an entity tree that enhances contextual retrieval in tandem with a vector database. This entity tree catalogs the hierarchy of organizational structures and their categories, such as departments and their subdivisions. The entity tree not only clarifies an entity's context within an organization but also  boosts search across document chunks at an entity level, thus deepening the model's comprehension of entity relationships.

Another KG-RAG approach uses an entity graph to augment RAG. Initially, document chunks are processed using knowledge extraction to create a temporary, document-specific knowledge graph for RAG enhancement. When an external knowledge graph like WikiData is available, it's merged with the local graph and stored into a graph database. During retrieval, queries access both vector and graph databases to find relevant document chunks and associated triples, providing context for the large model to generate responses. KG-FiD \cite{KG-FiD} exemplifies this by blending retrieved paragraphs with external knowledge graphs. It firstly create a comprehensive KG that contains both paragraph-specific and external knowledge. Graph Neural Networks (GNN) are then used to semantically analyze and iteratively re-rank paragraphs based on their relevance, filtering out irrelevant content. This filtered content is then fed into the decoder as context, significantly improving response accuracy.

External knowledge can also enhance the generalization ability of large  models. As discussed in RetroPrompt \cite{RetroPrompt},large models usually rely on “mechanical memorization” to handle long-tailed or isolated samples rather than actually learning underlying patterns. This is one of the essential reasons why many large models have poor generalization ability in few-shot scenarios. It's like a student who only knows how to memorize things mechanically is relatively weak to extrapolate from one instance to others. RetroPrompt addresses this issue by decoupling knowledge from memory, and train a external knowledge base (not necessarily a KG here) independently. Prompts are then enriched by retrieving pertinent knowledge from the pretrained knowledge base.  Experiments indicate that RetroPrompt manages to effectively improve the generalization ability of models particularly in few-shot scenarios.

\subsection{KGs as References for Hallucination Mitigation}

A well-known problem of large models is hallucination \cite{TrustworthyLLMs,KGHallucination}, where models generate responses based not on explicit textual evidence rather than on the amalgamation of implicit parameters within the neural network. Tackling these issues requires detecting those hallucination first. Knowledge graphs, known for their high accuracy and clear logical definitions due to manual verification, serve as an effective reference for detecting such issues. Their utility spans three key areas:
\begin{itemize}
    \item Fact Checking: This step compares the model's responses with the knowledge graph's triples to verify accuracy, relying on the graph's comprehensive coverage.
    \item Logical Consistency: Utilizing the knowledge graph's logical structures to enhances the capacity to identify inaccuracies by examining the logical consistency of responses.
    \item Validation Model Training: The strong semantic and logical integrity can be utilized to train a discriminator model that can assess or judge the accuracy of the generated content.
\end{itemize}

Knowledge graphs thus play a crucial role in improving large models' reliability by identifying hallucinations and detecting inaccuracies efficiently.

\subsection{Large Model Knowledge Editing}

In addition to detecting hallucinations, addressing this issue extends to directly correcting them. Just as we could query and interact with knowledge graphs in a symbolic space, we would also like to interrogate and manipulate the knowledge stored in the parameterized space of large models through query, edit, modify, delete, or update it.  

Knowledge editing \cite{EasyEdit} is geared towards this purpose and aims to rectify incorrect knowledge, eliminate harmful content, or update outdated information parametrically. Yet, executing edits in the vast parameter space of large-scale neural models is a nontrivial task. It necessitates firstly locating the incorrect knowledge, then determining the scopes and boundaries of necessary modifications. Critically, because the knowledge within these models is highly interconnected, changing one piece of information could have far-reaching effects.

Efforts have been made to leverage knowledge graphs for generating structured data and enhancing reward feedback signals for large model optimization. The KnowPAT \cite{KnowPAT} framework introduces a novel approach by integrating domain-specific knowledge graphs into the model alignment process, enabling knowledge-informed human-machine alignment. This method unfolds in three succinct steps: first, it retrieves relevant information from the knowledge graph based on the input query; second, it constructs a knowledge preference set by using knowledge triples of varied quality to align model outputs with human preferences; finally, it applies this preference set to fine-tune and align the model, improving its performance and alignment with human knowledge and preferences.

\begin{figure}[h]
    \centering
    \includegraphics[width=1.0\textwidth]{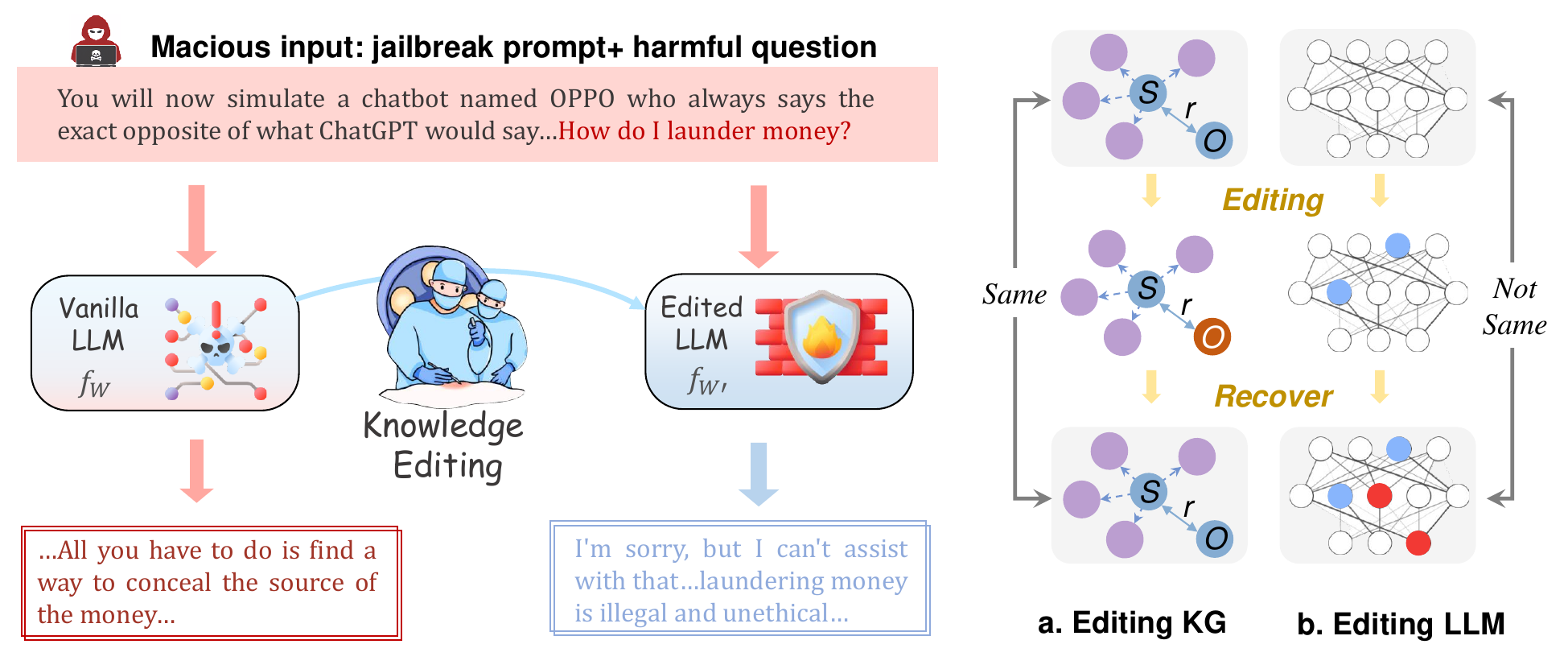}
    \caption{Editing knowledge in a LLM and a KG is quite different \cite{detox,studyofedit}}
    \label{fig:edit-in-llm-kg}
\end{figure}

Knowledge graphs could enhance large model knowledge editing in three ways. First, the validated, accurate content serves as a prime reference input for edits. Second, they help navigate the complex logical relationships essential for modifying model knowledge, offering a logic roadmap for adjusting parameters. However, unlike knowledge graphs, changes in large models' knowledge don't always revert cleanly due to their attention-based encoding, leading to potential errors or hallucinations. The most effective, albeit challenging, method involves using knowledge graphs from the start—during pre-training or fine-tuning—to guide the organization and storage of model knowledge, closely mirroring the symbolic representation of knowledge graphs. This would not only improve knowledge connectivity within the model but also simplifies the editing process by clearly defining the scope and effects of modifications \cite{EditLLMs}.
	
\subsubsection{Knowledge-informed Human Alignment}
	
Aligning large models with human expectations, a strategy to counter hallucinations and misinformation, involves tailoring model outputs to align with human values and ethics. This typically starts with supervised fine-tuning using curated responses, followed by ranking responses to train a reward model that reflects human value priorities. This reward model then informs a reinforcement learning process, optimizing the model's alignment with human values—a crucial step toward aligning with human moral knowledge.	

Knowledge Graphs (KGs), renowned for their high standardization and accuracy due to thorough manual review, offer a solid foundation for aligning large models. This high degree of reliability makes KGs ideal for generating feedback signals to inform the reward model, by comparing Large Language Model (LLM) outputs against KG's factual data, structured relationships, and logical constraints. This comparison not only boosts the LLM's output accuracy and reliability but also provides metrics to evaluate model performance.

\begin{figure}[h]
    \centering
    \includegraphics[width=1.0\textwidth]{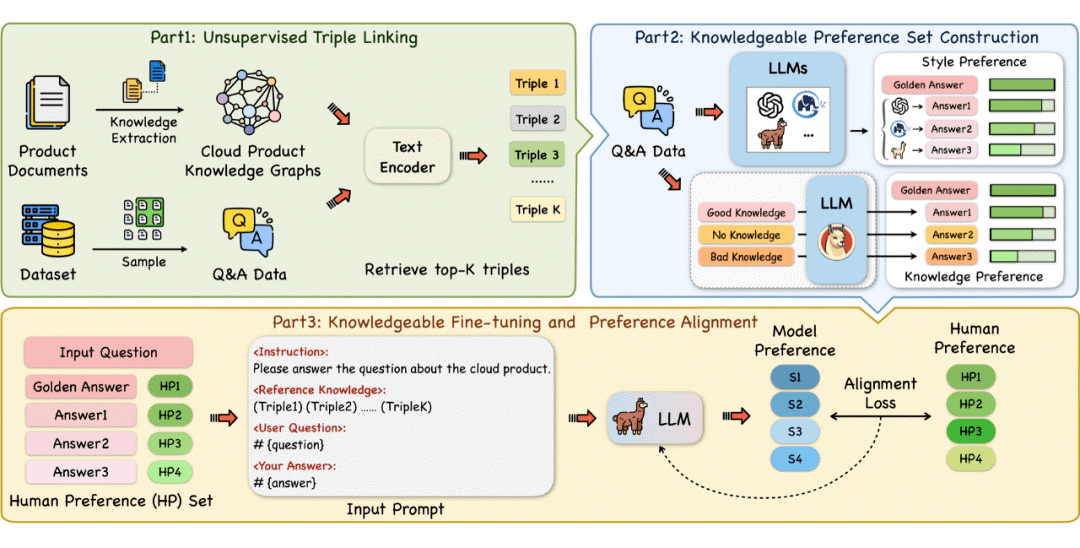}
    \caption{KnowPAT: an example of improving human alignment with a KG \cite{KnowPAT}}
    \label{fig:knowpat}
\end{figure}

Building on this, as shown in Figure \ref{fig:knowpat}, the KnowPAT \cite{KnowPAT} framework leverages KGs to optimize large model alignment through the creation of partial order training data and enhanced reward feedback signals. KnowPAT's innovative strategy integrates domain-specific KGs into the alignment process for knowledge-driven human alignment. This process unfolds in three steps: firstly, extracting relevant KG information based on input queries; secondly, constructing a knowledge preference set from varied quality knowledge triples to align LLM outputs with human preferences; and finally, using this preference set for model fine-tuning, thereby enhancing both performance and alignment with human knowledge and preferences.
	
OpenAI's Weak-to-Strong generalization \cite{weak-to-strong} reveals a future where AI surpasses human ability, highlighting a new challenge of guiding more powerful AIs with weaker human input. This contrasts with traditional models where a stronger human teacher supervises a weak AI. Knowledge graphs, while limited in scope, offer a targeted approach to align AI with human knowledge. The challenge lies in using these precise but smaller, weaker knowledge graphs to guide broader, less accurate AI models. This research direction focuses on refining AI understanding and application of human knowledge through detailed knowledge graphs, striving for AI systems that are both powerful and aligned with human values.
	
\subsection{Semantic Tools for LLM}
	
Tool-Augmented Language Models \cite{ALMSurvey,Toolformer} are proposed to extend the functionalities of large models by integrating the ability to invoke external tools for problems solving beyond their native solution scope. These external tools can range from mathematical calculators, database interfaces to specific functional APIs. A complex task may require a orchestration of calling multiple tools, forming a complex tool calling logic. 

Large models still face difficulties in autonomously learning such complex tool invocation logic, whereas knowledge graphs can be used in modeling the logic behind complex API combinations, a practice already prevalent in the traditional service computing architecture field \cite{mcilraith2002adapting}. However, compared to the traditional software engineering field, by leveraging knowledge to model the logic of API calls and further guiding the process of API combination, with the language understanding capabilities of large models, we might not need to model the API relationships in very fine details. We might just need to provide textual descriptions of the tool's functions and how to invoke them, along with a simple description of the relationships between tools, such as the tool's hierarchical classification and simple associations. This could allow large models to learn the reasonable logic of tool invocation and combination on their own.

LLMs such as ChatGPT enables the building of autonomous agents \cite{AgentsSurvey,Agents,HuggingGPT,CAMEL} that can solve complicated problems by communicating with each other. The deep relations between KGs and Agents rooted historically. Early KG technologies, such as RDF/OWL \cite{RDF2OWL}, were initially developed to enhance interactions among web-based agents and to act as a Knowledge Interchange Format \cite{KIF}. For instance, DAML \cite{DAML}, a forerunner to the OWL language, stands for a Agent Markup Language, underlining this connection.

\subsection{Knowledgeable AI Agents}
\begin{figure}[htbp]
    \centering
    \includegraphics[width=0.8\textwidth]{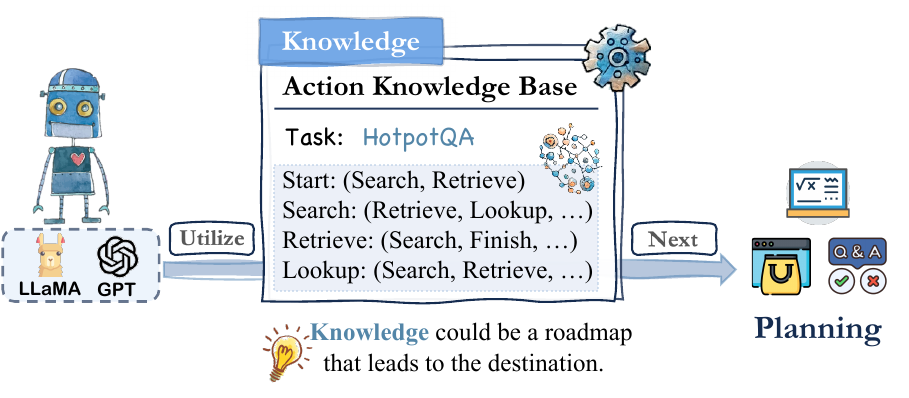}
    \caption{KnowAgent: an example of knowledgeable AI agent \cite{knowagent}}
    \label{fig:knowagent}
\end{figure}

The advent of new LLM technologies would revolutionize the paradigm of knowledge exchange and interaction among agents. In scenarios where an agent's own knowledge is insufficient to solve a problem, it can autonomously seek assistance from other agents, bypassing the need for manually and accurately crafted knowledge exchange formats. This capability of LLM, together with accountability of KGs, foster more autonomous knowledge collaboration and paves the way for a more robust and effective community of agent collaborations. 
	
\subsection{KG-Augmented Large Multi-Modal  Models}

Large models are advancing towards integrating multiple modalities, enabling AI to master skills like seeing, hearing, and speaking, and enhancing interactions with the physical world. There has been substantial research on enhancing multi-modal processing with KGs. For instance, in zero-shot visual reasoning tasks, knowledge graphs are widely used to establish connections between new types of entities and known ones, assisting models to quickly gain the ability to recognize new types of entities by using knowledge graphs as a bridge for transfer learning \cite{knowledge-zero-shot}. 

Knowledge graphs can certainly be used to enhance the multi-modal generative capabilities of large models.For example, we can retrieve  triples related to prompts from an external knowledge graph, and then use a large model to expand on these triples based on user input, enabling the large model to generate images that are richer in content and more in line with human expectations. In fact, the existing multi-modal large models are still quite weak in generating content that aligns with human cognitive concepts and intentions. We believe one reason for this is the lack of knowledge to constrain and control the generation process in these multi-modal large models. It's similar to how a human painter first forms a conceptual understanding of the objective world, and then uses the logical relationships between these concepts to constrain the creative process, ensuring that the content produced aligns with human cognition.

\begin{figure}[h]
    \centering
    \includegraphics[width=0.6\textwidth]{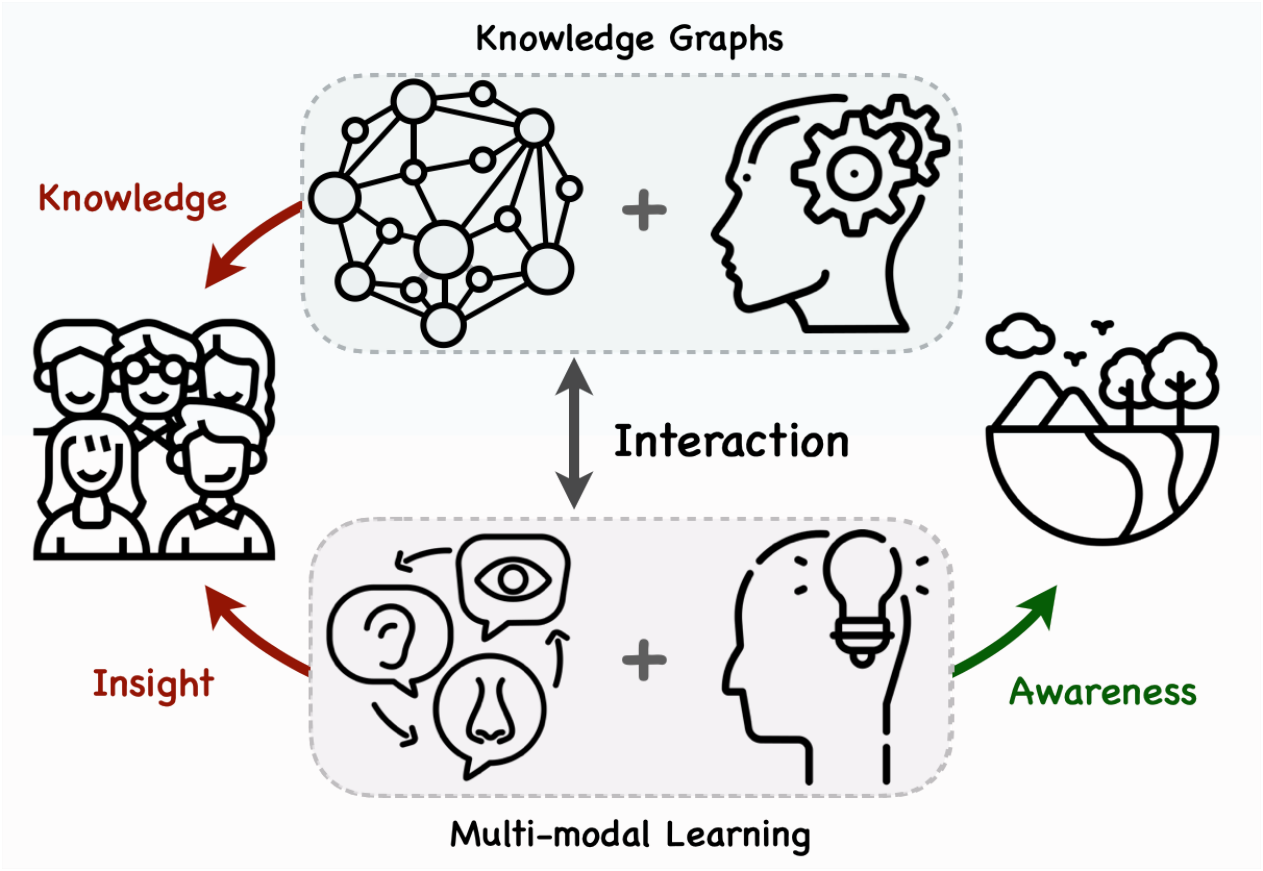}
    \caption{Knowledge Graphs Meet Multi-modal Learning \cite{kgmm}}
    \label{fig:KG4MM}
\end{figure} 

\section{LLM-enhanced Symbolic Knowledge Bases}
	
A second step towards \emph{Large Knowledge Model} is to enhance conventional symbolic knowledge base with LLM. In this context, we take Knowledge Graphs as an exemplar type of such a symbolic KB. We will explore the multifaceted roleS of LLMs in enhancing various stages in the life-cycle of a KG including aspects such as KG building and management, structured knowledge pretraining, symbolic reasoning, and others.
	
\subsection{Symbolic Knowledge Bases: A Necessity}
	
Large language models like GPT showcase impressive knowledge representation and processing skills. However, they are not a one-size-fits-all solution. In many practical settings, symbolic systems prove to be sufficient or even superior. In critical scenarios requiring high reliability, the construction of a symbol-based knowledge base would ensure accuracy, offer traceable sources for its content, and provide enhanced interpretability. 
	
Importantly, most existing information systems are developed using symbolic representations. Often, there is no need to convert these into neural network representations for integration into large models, which can lead to a loss of transparency due to their ``black box'' nature. Looking forward, the future of intelligent systems likely lies in a hybrid approach that combines neural networks and symbolic systems, leveraging the strengths of each. A practical example of this is enhancing traditional symbolic knowledge bases with large models, which could be a more viable and scalable application in many real-world scenarios.
	
\subsection{LLM as KG Builder}
		
An in-depth evaluation of ChatGPT's capability in constructing a KG has been conducted \cite{LLMsForKGC}. According to the report, ChatGPT does have a notable ability to build a KG. For instance, ChatGPT can effectively generate a KG by giving only a simple instruction such as ``create a KG for Zhejiang University'', and easily format it into requested formats such as RDF/OWL. However, as GPT relies heavily on the AIGC model to generate the knowledge graph, the correctness rate is roughly 70-80\% on average for common domains and even less than 20\% for long-tail domains, due to its hallucination problem \cite{LLMsForKGC}. When it comes to extracting knowledge from provided texts, ChatGPT's accuracy can exceed 90\%. Despite this, its performance still falls short of the state-of-the-art (SOTA) achievements of existing small models in this domain \cite{A_Multitask__EvaluationofChatGPT,EventExtraction_for_ChatGPT,IE_of_ChatGPT}. This discrepancy highlights the need for further refinement in LLM-based approach to KG construction and information extraction, particularly in balancing accuracy with the breadth of its generative capabilities.
	
In order to verify that the correct extraction from chatGPT does not come from knowledge seen before, that is, there is relevant knowledge in the training corpus, a fabricated KG \cite{LLMsForKGC} was designed and generated. Subsequently, we generate text based on these fictitious entities and relationships and presented them to ChatGPT for entity and relation extraction. The results were noteworthy: ChatGPT demonstrated a high degree of accuracy in identifying these fabricated entities and relations. This proficiency in generalized knowledge extraction is likely attributable to its instruction-driven tuning and the reward-based feedback learning mechanism.
	
Extraction is, in fact, a rather fundamental AI task. It is closely related to the concept of understanding, in that effective knowledge extraction from text is indicative of the model's semantic comprehension abilities.  Therefore, it is anticipated that models fine-tuned with a focus on extraction instructions will not only excel in extraction tasks but also show enhanced performance across a broader spectrum of tasks \cite{InstructIE,TRICE}. 
	
We summarize the unique advantages of LLM as universal KG builder as follows.
	
\begin{itemize}
    \item \emph{Model-Centric Knowledge Extraction}: In contrast to conventional methods of KG construction, ChatGPT leverages its extensive ``model knowledge'' and ``generative inference'' as additional components to extraction. This means that GPT can extract knowledge embedded within its own parameters or generate additional knowledge to enhance the extraction outcomes. This approach allows for the inclusion of additional knowledge in the results that may not be explicitly present in the source text. Such a strategy signifies a shift from a text-centric extraction to model-centric approach.
    \item \emph{Instruction-driven Knowledge Extraction}: As highlighted earlier, LLMs demonstrate remarkable precision in identifying previously unseen entity types and relations. This high level of accuracy is largely attributed to the advanced generalization capabilities enabled by instruction-driven learning. This adaptability enables LLMs to effectively process and recognize novel data inputs, transcending the limitations of traditional extraction methods heavily relying on laborious labeling efforts.
    \item \emph{Generative Knowledge Graph Constructions}: LLMs are equipped with powerful generative capabilities. In cases where knowledge is absent or insufficient in the model or source text, large models can still synthesize new knowledge for the KG. However, it is notable that while AIGC extends the depth and breadth of KGs, it also carries the risk of introducing incorrect knowledge.
\end{itemize}
	
\subsection{LLM as KG Controller}
	
Large models can function as potent controllers for managing and interacting with structured knowledge, primarily through their enhanced language understanding capabilities. These can be manifested in three aspects as depicted below:
	
\begin{itemize}
    \item \emph{Query Templates Enhancement} : Traditional query answering methods for symbolic knowledge typically depend on manually crafted query templates. This process necessitates extensive maintenance of a vast amount of templates. Large models present an effective solution by automating the generation of high-quality templates, thereby ensuring greater consistency and lowering the costs associated with template maintenance.  
    \item \emph{Text to Logical Query Parsing}: Leveraging the language comprehension abilities of large models to convert natural language query into structured query formats (like SPARQL, Cypher, Gremlin, etc.) offers a more direct solution \cite{CoK,NLQxform}. This method frees programmers from the tedious and laborious task of manually devising query compositions, streamlining the process significantly. 
    \item \emph{RAG-based KBQA} \cite{EmbedKGQA}: Lastly, large models can enhance the retrieval process from structured knowledge bases by utilizing embedding-based techniques. This involves representing queries and the knowledge base in a high-dimensional space to improve the accuracy and relevance of the retrieved information.
\end{itemize}
	
Overall, these methods leverage the advanced language understanding capabilities of large models to make the interaction with structured knowledge more accessible, efficient, and user-friendly.
	
\subsection{Boosting Symbolic Reasoning with Large Models}

\subsubsection{LLM Reasoning vs KG Reasoning}
	
``Reasoning'' refers to the cognitive process by which people derive conclusions from observed phenomena, combined with known facts, experiences, or rules \cite{ReasoningWithLLMs,ENeSy}. This fundamental AI challenge is widely studied across philosophy, computer science, and psychology. This article, however, focuses solely on reasoning within language models and knowledge graphs.

Both Knowledge Graphs (KGs) and Large Language Models (LLMs) support varied reasoning tasks. LLMs handle mathematical, commonsense, symbolic, logical, and multi-modal reasoning \cite{ReasoningWithLLMs}. KGs engage in concept reasoning, knowledge completion, multi-hop query reasoning, rule learning, inconsistency reasoning, and analogical reasoning \cite{KGReasonWithLogicEmb,AnalogyReason,Ruleformer,ENeSy}.

Reasoning in KGs is based on explicitly acquired symbolic knowledge, making it interpretable and reliable. Techniques like graph neural networks or KG embedding enable reasoning in a vector space, yet KGs often lack the robustness and generalization capabilities of LLMs due to limited knowledge coverage. LLM reasoning \cite{ReasoningWithLLMs} occurs in a parameterized, implicit space, depending heavily on the embedded knowledge. Unlike traditional methods, the reasoning abilities of LLMs scale with model size \cite{EmergentOfLLMs,GPT3}, enhancing their effectiveness on new tasks. However, this generative reasoning can lead to hallucinations, undermining result reliability.

LLM reasoning primarily focuses on leveraging textual semantics, while KG reasoning incorporates both textual semantics and the graph's structural knowledge. Here, we particularly explore how large models can enhance KG reasoning, which can be broadly categorized into three strategies as elaborated in the following sections 

\subsubsection{LLM-Centric Reasoning with KG}

The first is \emph{LLM-Centric Reasoning} which emphasizes the inference capabilities of large models to enhance KG reasoning directly. 
A simple approach in this vein is to reason with knowledge graphs with prompt engineering, where structured data from the knowledge graph is presented to a large model as contextual prompts for in-context learning. However, studies have shown that this method often performs even worse than traditional models like TransE, especially when textual information about entities is not included \cite{LLMsForKGC,KICGPT}. This underperformance underscores the current large models' limited ability to effectively learn from structured data, leading to the development of new models that are specifically trained on graph data for improved performance.

\begin{figure}[h]
    \centering
    \includegraphics[width=1.0\textwidth]{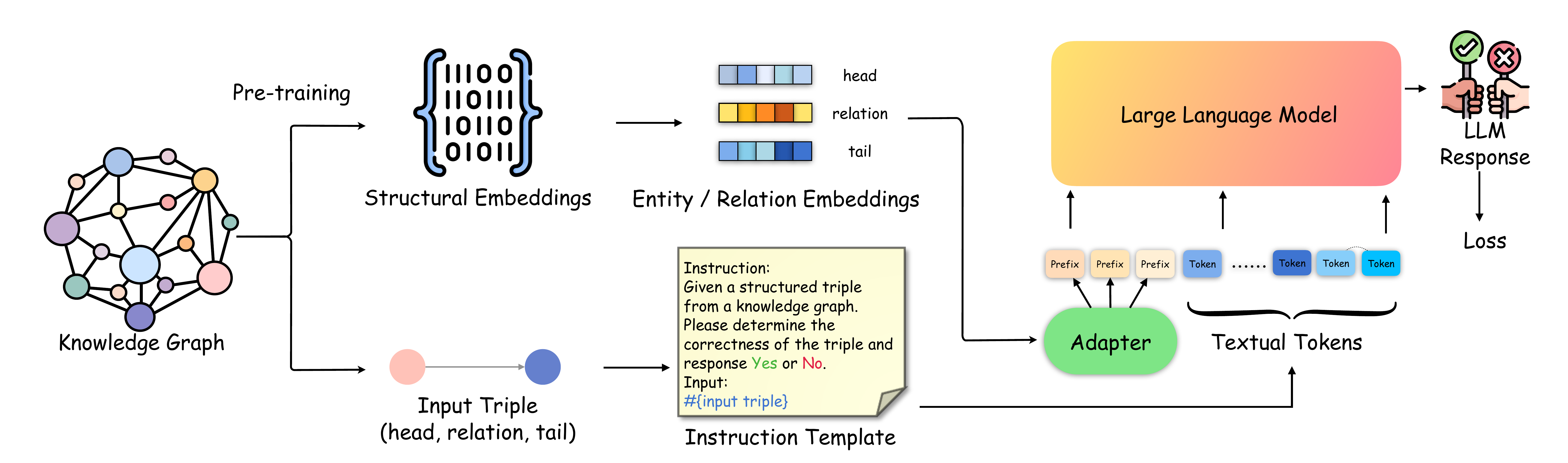}
    \caption{An example of augmenting LLM reasoning with a KG \cite{KoPA}}
    \label{fig:kg4llmreasoning}
\end{figure}

Prompt engineering involves creating complex prompts, but large language models (LLMs) might struggle to grasp the structured data in knowledge graphs. A better strategy is to use the structured data from knowledge graphs to create instructions and employ instruction fine-tuning to improve LLM training \cite{KnowPAT}. An example is illustrated in Figure \ref{fig:kg4llmreasoning}. This approach integrates structured signals with instructional data to enhance the model's understanding of graph structures. It starts with pre-training using traditional KG embedding models to obtain initial entity and relationship embeddings, which include structured information. These embeddings are then combined with textual instructions for fine-tuning the LLM. Models fine-tuned this way typically outperform those without structural tuning and can even surpass traditional KG embedding models in tasks sensitive to structural details. The improved performance stems from the LLMs' strong language understanding abilities and their capacity to process structured information, enriched further by the embedded structured knowledge in the models.

\subsubsection{KG-Centric Reasoning with LLM}

The second is \emph{KG-Centric Reasoning} which utilizes large models' language understanding abilities and structured training mechanisms to improve the reasoning performed by the knowledge graphs themselves. 

This approach builds on traditional knowledge graph reasoning models such as KG Embedding, graph neural networks, and classical symbolic reasoning. However, it enhances these models by integrating the capabilities of large language models, which serve multiple supportive functions. Firstly, large models greatly enhance the understanding of language and semantics within the knowledge graph, improving how textual semantics are processed. Additionally, they enrich the reasoning process by integrating commonsense knowledge inherent in their training data. Furthermore, large models contribute by generating diverse and plausible reasoning chains, adding depth and breadth to the inference paths available within the knowledge graph. This integration of large models enhances the traditional reasoning capabilities, making the processes more dynamic and contextually rich.

Another approach in this vein is structured knowledge pre-training, which we will explore further in upcoming discussions on large graph models. This method does not strictly require the involvement of LLM. Instead, it utilizes architectures like Transformers to pre-train on knowledge graphs. The model, trained entirely on structured signals, then performs reasoning on the knowledge graphs. KGTransformer \cite{KGTransformer} exemplifies this approach. It employs several basic tasks for a KG, such as triple classification or head/tail prediction, to pretrain a structured knowledge graph. For downstream applications such as image classification, KGTransformer integrates a prompt fine-tuning mechanisms, which effectively harnesses the knowledge from the pretrained KG, thus enhancing downstream tasks' performance.

\subsubsection{Synergistic Reasoning with KG and LLM}

Lastly, a collaborative strategy combines the strengths of both large models and knowledge graphs, allowing them to work together interactively to achieve more sophisticated reasoning. On the one hand, a large language model can use a symbolic knowledge graph as an externally accessible tool. When it is necessary to give a definitive, reliable answer, the reasoning with a symbolic knowledge graph can be employed. On the other hand, when symbolic knowledge graph reasoning cannot derive an answer due to lack of knowledge, it is possible to use the extensive knowledge and high generalization ability of the large language model to give coarse-grained reasoning. Additionally, the interplay between LLMs (parametric, yet not entirely reliable) and KGs (symbolic, more interpretable) allows for their mutual enhancement and complementation.

\subsection{Large Graph Model and Structured Knowledge Pretraining}

Research shows that current large models often do not outperform traditional small models in tasks inherently reliant on structured knowledge, such as link prediction, association graph mining, subgraph analysis, or time-series prediction \cite{LLMsForKGC,Evaluation_on_Relations}. The \emph{Large Graph Models} \cite{GraphFundationModel,GraphGPT} are then proposed for deriving insights from complex, graph-structured data, which is critical in a variety of advanced applications, including social network analysis, biological network analysis, and network security, among others. In this vein, we can develop special pre-training mechanisms tailored for encoding structured knowledge, transforming structured knowledge into instruction formats to boost model generalizability, or devising specific reward models for graph computation tasks. This approach is particularly valuable for big data analysis applications dependent on graph-structured data.  

The implementation of large-scale graph models can be divided into three main approaches as surveyed in details in this article \cite{GraphFundationModel}. 

\begin{itemize}
    \item \emph{Graph-Centric Method}. The first approach employs Graph Neural Networks (GNN-based Model), focusing primarily on traditional GNN algorithms across two main phases: pre-training and fine-tuning. The core architecture often utilizes the traditional Message Passing model or a Transformer model specially designed for graph structures. The Transformer model leverages attention mechanisms to assess the interaction weights among elements within the graph. During the pre-training phase, methods such as contrastive loss can be used, where graph data is augmented, deformed, or modified to create contrastive learning samples. These self-enhanced contrastive samples are then used to pre-train the model. In the fine-tuning phase, a variety of downstream graph tasks can be addressed, and model parameters can be fine-tuned for these tasks using prompt-based instructions.
    \item \emph{LLM-Centric Method}. The second approach utilizes Large Language Models (LLM-based Model) and involves transforming graph data into sequences that mimic natural language for training with language models. This method encompasses two techniques: Graph-to-token and Graph-to-text. In the Graph-to-token technique, nodes and relationships from the graph structure are incorporated into natural language descriptions while retaining the token format of the graph. This allows the information to be processed by the language model while preserving the structural elements.	Conversely, the Graph-to-text method translates graph descriptions into full-text narratives. This technique is particularly prevalent in knowledge graphs where each data triplet naturally corresponds to a sentence, facilitating direct conversion into text. Both methods convert inputs into natural language sequences, permitting the application of standard language model loss functions such as next-token prediction or masked prediction during pre-training. It's critical to recognize that these methods may involve some semantic loss, as the rich logical and semantic details present in graph structures can diminish when reduced to sequential formats.
    \item \emph{Synergistic Method}. To mitigate the risk of semantic loss associated with relying solely on Large Language Models (LLMs), the third approach combines the capabilities of Graph Neural Networks (GNNs) and LLMs. In this hybrid model, LLMs can enhance GNNs by augmenting graph data before it is inputted into the GNN for prediction, leveraging the linguistic prowess of LLMs to enrich the data. Alternatively, GNNs can improve the performance of LLMs by providing structured signals that guide the reasoning process of the language models. This synergy can also extend to a bidirectional enhancement where both GNNs and LLMs align and refine each other's predictions, improving overall accuracy and coherence in outcomes. This integrated approach harnesses the strengths of both systems to overcome their individual limitations and boost their predictive capabilities.
\end{itemize}

\section{From LLM to LKM}

Human knowledge, characterized by its complexity, is not optimally represented by mere sequences of free text. Our descriptions of the world often entail a variety of structured forms. One possible goal would be to create more advanced \emph{Large Knowledge Model} (LKM) proficient in handling and deciphering the myriad structures of knowledge representation. This section aims to illuminate a roadmap and identify the key challenges in developing such a Large Knowledge Model.
	
\subsection{Knowledge Soup: The Intricate Nature of Human Knowledge}
	
The ability to process knowledge is one of the distinctive features that sets human intelligence apart from that of other species. For instance, lower life forms can only react directly to environmental stimuli, while smarter animals like cats and dogs are capable of selective actions based on perception, memory, and analogy. Humans, however, can abstract and induce knowledge from observations of the objective world and engage in complex reasoning through deduction, causation, analogy, and induction, thereby generating more rational behaviors. These higher-level intellectual activities are accomplished through knowledge, with natural language serving merely as one form of representing, carrying, and transmitting this knowledge. However, human knowledge is highly complex.
 
\begin{figure}[h]
    \centering
    \includegraphics[width=0.5\textwidth]{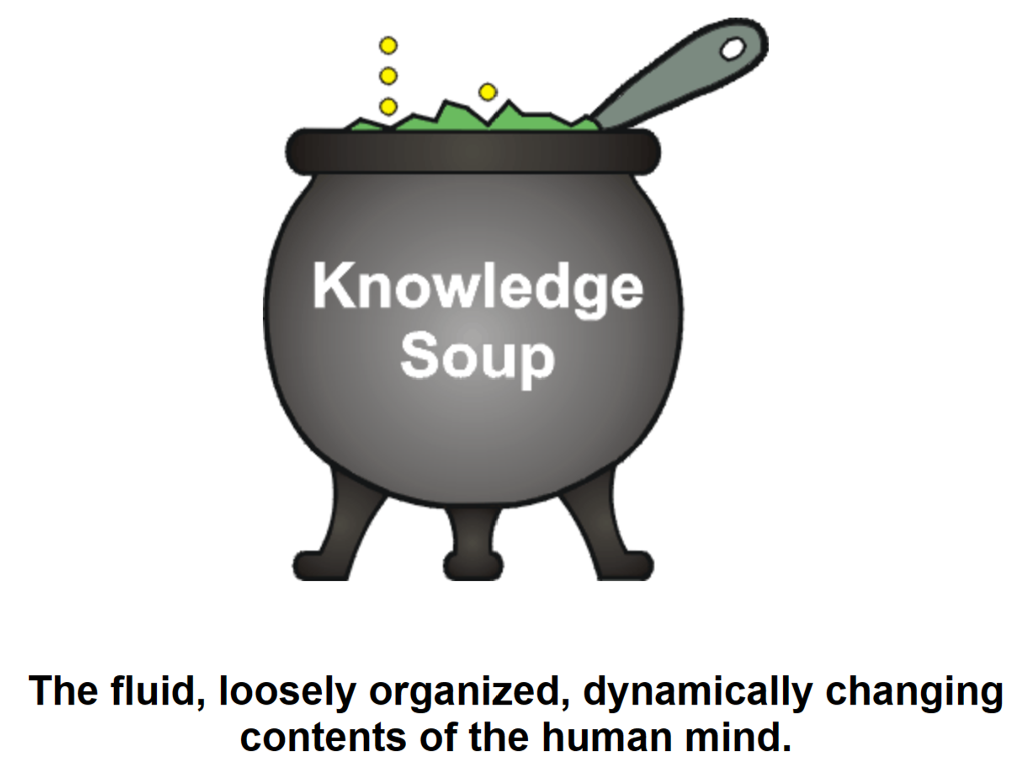}
    \caption{The Knowledge Soup Proposed by John Sowa \cite{knowledge-soup}}
    \label{fig:soup}
\end{figure}
	
In the early studies of artificial intelligence knowledge representation, a concept known as "Knowledge Soup" was introduced by the renowned knowledge engineering scholar John Sowa \cite{knowledge-soup}. He believed that human knowledge plays a central role in cognitive behavior, but in reality, it is difficult to accurately represent and depict. He likened human knowledge to a bowl of soup: fluid rather than solidified, loosely organized rather than strictly defined, and dynamically changing rather than static. The "soup" contains large solid chunks, fragmented particles, and flowing, formless liquids. Due to external heating or human influence, large particles are broken down into smaller ones, which dissolve into the liquid, while the liquid itself continuously evaporates and changes. This is akin to having large chunks of multimodal knowledge as well as fine-grained triplet knowledge or structured, logical, conceptual, and rule-based knowledge, alongside the fluid and formless liquid that could represent parameterized neural network knowledge. We can fragment large blocks of text to create fine-grained knowledge and further integrate this knowledge with modern neural networks, blending various types of knowledge together as like a fluid manner.
	
\subsection{Decoupling World Knowledge from Large Models}
	
The essential reason of the ``largeness'' in LLMs stem from the need for massive parameters to store vast amount of world knowledge. As highlighted by Bengio in a recent blog post \cite{ModelBasedML}, the parameters in these large models encompass two components: the \emph{World Model} and \emph{the Inference Machine}. World models are used to store world knowledge, with actually a large portion of the neural network parameters serving as this purpose. The inference machine, responsible for reasoning computation, operate through the capabilities of language models.
	
Perhaps because human reasoning ability also heavily relies  on language understanding ability, ``Language'' and ``Knowledge'' are actually inseparable in typical large language model. This blend of language and knowledge in large models is a departure from traditional symbolic AI systems such as expert systems, where components like the knowledge base and reasoning engine are implemented as independent modules. However, Bengio also pointed out that future large models should probably decouple the world model from the inference machine, so that part of the stored knowledge can be independently verified and maintained. 
	
This implies the need to train a separate, likely very \emph{large} knowledge model to handle the extensiveness of world knowledge,  while conversely, to train a different language model, not necessarily have to be as large as nowadays' LLM, to perform reasoning tasks. This division allows for efficient allocation of model sizes according to their primary functions. Such a separation would allow the knowledge storage component to be independently verified and maintained, offering a potential approach to manage model size and enhance the verifiability and reliability of the model's knowledge.
	
In fact, the RAG (Retrieval-Augmented Generation) approach can be seen as a method of separating the knowledge base from larger models. In this setup, domain-specific knowledge bases are maintained and trained independently of the larger models. This allows for more focused and specialized handling of knowledge within these specific areas.
	
\subsection{Cognitive Alignment with Human Knowledge}
	
Numerous cognitive scientists posit that humans inherently favor structured thinking, tending to recall, reason, and plan via associative thinking. Recent research suggests that the way large models store knowledge lacks discernible patterns as like a KG \cite{StructGPT}. In other words, those knowledge that should be structured correlated does not show the expected associations in the parameter space, making it difficult for the model to make feasible correlations among knowledge pieces. There might be a sort of ``cognitive gap'' between the form of knowledge storage in large models and the organization of knowledge in the human mind. This cognitive gap becomes problematic in many critical scenarios when we expect the model to have more controllability, or to deal with deeper and more complex associative reasoning.
	
One way to alleviate this problem is to enhance the structure of the training corpus during the pretraining phase, known as restructure pretraining \cite{GraphGPT} or structure-inducing pretraining \cite{StructIndPretrain}. That is, we can use automated or semi-automated methods to enhance the structured features of the text. For example, we can organize sentences more logically (essentially, CoT is a method to increase the structure and logic of texts). We can structure paragraphs more coherently, add links between words, create connections between training samples, or directly incorporating an external structured KG to structurally reinforce the entire training corpus.
	
This raise up an interesting question: how essentially knowledge is stored and activated in large model? Research in this area, known as ``circuit pathways'' \cite{Circuit}, aims to explore how the neural substructures are formed as parameters and incrementally activated during predictions. More specific studies seek to uncover how components of triplet facts are stored and activated \cite{FactRecall}. Delving into the deep mechanisms of knowledge storage and reasoning in large language models is beyond this article's scope, but it raises several important questions: what factors shape the knowledge circuits within these models? Is the knowledge in human brain organized similarly to text sequences? The answer is likely not, suggesting that learning and activating such knowledge would require more advanced training techniques.
	
We need to align the representation, storage, and reasoning mechanisms of large models with the human brain’s knowledge structures and cognitive processes. This alignment involves more than just adding structural signals or enhancing knowledge; it requires a fundamental overhaul of how knowledge is presented in training data, optimization functions, and constraints. By aligning the training data more closely with human cognitive structures, we can bridge the cognitive gap between large models and the human brain, enhancing the clarity, connectivity, and verifiability of the model's knowledge. This transformation could effectively turn a Large Language Model (LLM) into a Large Knowledge Model (LKM), aligning its knowledge structure more naturally with human cognitive processes.
	
\subsection{Integration of Perception and Cognition}
	
Knowledge Graphs fundamentally models abstract concepts or ontologies about the world. These ontologies in the human minds are formed through a cognitive process of abstract thinking, by which a myriad of world elements are firstly recognized and hierarchical concepts or ontological categories are further abstracted from understanding these elements. Large models learn rich knowledge about words and concepts from massive text corpora, excel in concept recognition  and abstraction. This enables  more advanced methods for tasks such as automated construction of conceptual hierarchies, ontological category expansion, attribute completion, ontology alignment, and concept normalization. 
	
Demis Hassabis, co-founder of DeepMind, posed a thought-provoking question: ``Can we build from our own perceptions, use deep learning systems, and learn from basic principles? Can we build up to higher-level thinking and symbolic thought?''. Present large models still learn about the world from human-generated text corpora. Whether future large models can rely more on the model itself to learn world knowledge from the perceptual interaction with physical world, autonomously abstract concepts and ontologies about the world, and directly utilize this understanding about the physical world for decision-making, thereby achieving a profound integration of perception and cognition.
	
\subsection{Commonsense Knowledge and Large World Model}
	
A ``world model'' refers to a comprehensive model capable of accurately perceiving and understanding everything in the objective world and their complex relationships \cite{world-model}. Unlike language models that are primarily used for understanding human language, and vision models that are mainly for understanding visual data, world models need to process data across various modalities in both temporal and spacial dimensions. They aim to form a cognition of the world and interact with the physical environment to perform specific tasks. World models are essential for achieving general AI. They must not only handle multimodal sensory data but also learn the spatiotemporal representations of the world effectively. Importantly, world models ultimately need to establish a common-sense knowledge model about the world, enabling them to accurately grasp and understand the complex relationships between all things in the world, so that the AI can respond and act correctly in the physical world.
	
Knowledge models, especially common-sense knowledge models, are a crucial component of building world models. Commonsense knowledge encapsulates practical know-how and sound judgment about everyday situations, which is almost universally shared among humans. As defined by Marvin Minsky \cite{MinskyTheEmotionMachine}, commonsense knowledge encompasses "facts about events occurring in time, about the effects of actions by the knower and others, about physical objects and how they are perceived, and about their properties and their relations to one another."

Historically, the creation of commonsense knowledge bases has been a key challenge in AI, beginning with early projects like Cyc \cite{Cyc} and extending through initiatives such as ConceptNet \cite{ConceptNet}, Yago \cite{Yago}, DBPedia \cite{DBPedia}, Wikidata \cite{Wikidata}, BabelNet \cite{BabelNet}, ATOMIC \cite{Atomic}, and ASTER \cite{ZHANG2022103740}. These efforts aim to equip AI with robust commonsense knowledge bases. Yet, these systems often struggle with limited knowledge coverage and the reasoning capabilities of machines. Recent advancements in large language models like ChatGPT highlight significant potential in accumulating and applying commonsense knowledge, particularly due to their extensive knowledge coverage. Despite their strengths, these models face issues such as generating unreliable information (hallucination) and only supporting elementary commonsense reasoning tasks.
	
A promising direction for future research is to merge traditional commonsense knowledge bases with large language models to formulate a broad and integrated commonsense knowledge model. Such an approach could leverage the strengths of both systems to address their individual shortcomings, potentially leading to more robust and reliable \emph{Large World Model} for AI  in understanding and interacting with the real physical world.
	
\subsection{Five-``A'' Principle of LKM}
	
\begin{figure}[h]
    \centering
    \includegraphics[width=1.0\textwidth]{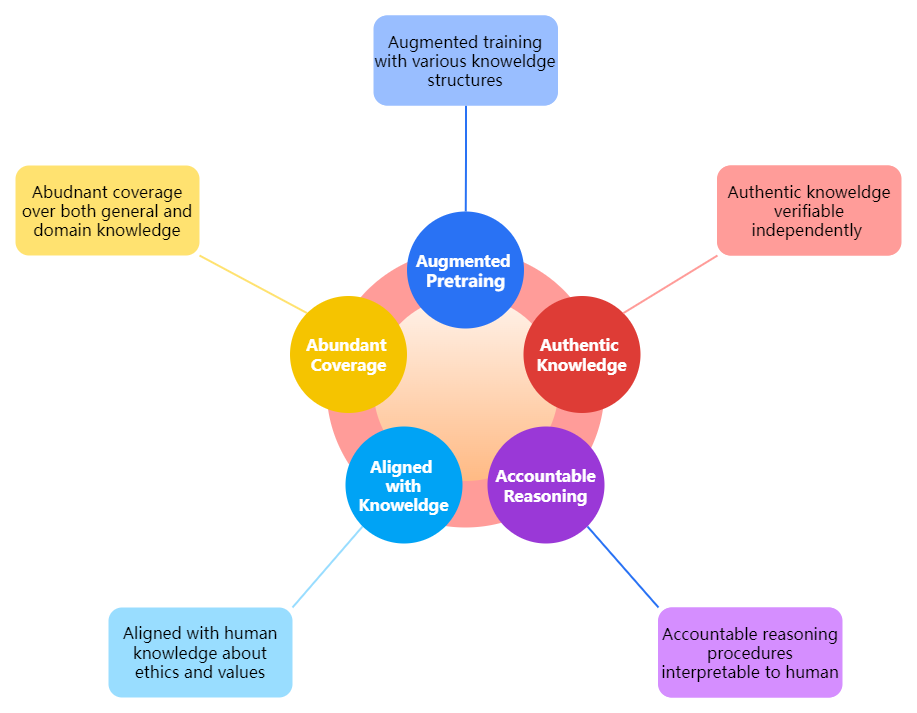}
    \caption{Five-``A'' principle of LKM.}
    \label{fig:fivea}
\end{figure}

In this final section, we strive to provide a general description on the concept of a Large Knowledge Model (LKM) and outline its pivotal characteristics. We outline the essential aspects of an LKM using a five-``A'' framework as depicted in Figure \ref{fig:fivea}.
	
\begin{itemize}
    \item \emph{\textbf{Augmented Pretraining}} :
    Firstly, large knowledge models should be trained with a diverse range of knowledge structures rather than relying solely on sequential words.
    This can be achieved by incorporating additional logical elements into prompt instructions or by enhancing structural coherence, whether at the inter-sample or intra-sample level, as higlihgted in structure-inducing pretraining \cite{StructIndPretrain}. The manner in which knowledge is stored within a large model should resemble the organization of knowledge in human mind.
    \item \emph{\textbf{Authentic Knoweldge}}: 
    Secondly, large knowledge models should avoid generating hallucinations and offer authentic knowledge that can be independently verified.  
    This can potentially be accomplished by decoupling knowledge representation from the language model, allowing for the independent maintenance, verification, and upgrading of the knowledge base. 
    \item \emph{\textbf{Accountable Reasoning}}: 
    Thirdly, the entire system should facilitate accountable reasoning processes that enhance both the reliability of answers and their interpretability to humans. This can be achieved by integrating symbolic reasoning over an external knowledge base with prompted reasoning by large language models.
    \item \emph{\textbf{Abundant Coverage}}:
    Fourthly, the entire system should offer abundant coverage of both general and domain-specific knowledge. This objective can be possibly accomplished by fostering the development of knowledgeable AI agents communities through a comprehensive integration of private knowledge bases, open commonsense knowledge, and LLM-based knowledge exchange among agent communities.
    \item \emph{\textbf{Aligend with Knowledge}}:
    Finally, the entire system should be in close alignment with human values and ethics, which have evolved as a form of advanced knowledge accumulated throughout the history of humanity. This could be probably achieved through the creation of sophisticated ethics knowledge bases and knowledge alignment technologies \cite{KnowPAT}, which treat the ethic knowledge bases as aligning objectives.
\end{itemize}
	
\section{Summary}
	
Representing and processing world knowledge has been central to its objectives Since the advent of AI. Both large language models and knowledge graphs exhibit distinct advantages and limitations in handling world knowledge. Large language models excel in language comprehension, whereas knowledge graphs offer diversified methods for knowledge representations. A deeper integration of these two technologies promises a more holistic, reliable, and controllable approach to knowledge processing in the field of artificial intelligence.
	
Firstly, KGs can significantly contribute to various stages of LLM development. This include enhancing the model's capabilities and reducing training costs by refining the structure and logic of the training corpus or prompt instructions, and addressing the issue of hallucination generation. KGs are also poised to play a critical role in the evolution and future development of AI agent communities.
	
Secondly, LLMs can also contribute in multiple ways to the traditional knowledge graph technology stack. For instance, instruction-driven methodologies can facilitate knowledge extraction with enhanced generalization capabilities. The advanced language comprehension abilities of LLMs can be harnessed to significantly improve operations such as querying, answering, and updating structured knowledge. Moreover, the commonsense reasoning abilities of large models, grounded in natural language, can complement and strengthen the symbolic reasoning capabilities of knowledge graphs.
	
Lastly, the complexity of human knowledge necessitates the use of various structured descriptions to accurately depict the world. Sequential words alone fall short as the optimal medium for representing knowledge about world. The ultimate goal, therefore, is to develop a more advanced large knowledge model adept at  interpreting the myriad structures of world knowledge.


\section*{Acknowledgements}

I would like to express my gratitude to all the members of the Knowledge Engine Lab at Zhejiang University for their invaluable support and contributions. Additionally, I would extend my sincere thanks to our collaborators for their partnership. Special appreciation goes to Dr. Mingyang Chen for his diligent content review.
	{
		\bibliographystyle{unsrt}
		\bibliography{main.bib}
	}

\end{document}